\documentclass[journal,12p]{IEEEtran}

\usepackage{amssymb}
\usepackage{latexsym}

\usepackage{algorithm}
\usepackage{algorithmic}
\usepackage{mathtools}
\usepackage{url}
\usepackage{xcolor}
\definecolor{newcolor}{rgb}{.8,.349,.1}
\usepackage{epsfig}
\usepackage[T1]{fontenc} 
\usepackage[utf8]{inputenc}

\begin{document}

\title{On the Information Content of Predictions in Word Analogy Tests}
\author{Jugurta~Montalv\~ao
\IEEEcompsocitemizethanks{\IEEEcompsocthanksitem J. Montalv\~ao is with the Department
of Electrical Engineering, Federal University of Sergipe, S\~ao Crist\'ov\~ao, 49000-000.\protect\\
E-mail: jmontalvao@academico.ufs.br
}
\thanks{Digital Object Identifier: 10.14209/jcis.2022.18}}


\setcounter{page}{175}
\markboth{JOURNAL OF COMMUNICATION AND INFORMATION SYSTEMS, VOL. 37, No.1, 2022.}{}

\IEEEtitleabstractindextext{%
\begin{abstract}
An approach is proposed to quantify, in bits of information, the actual relevance of analogies in analogy tests. The main component of this approach is a {\em soft} accuracy estimator that also yields entropy estimates with compensated biases. Experimental results obtained with pre-trained GloVe 300-D vectors and two public analogy test sets show that proximity hints are much more relevant than analogies in analogy tests, from an information content perspective. Accordingly, a simple word embedding model is used to predict that analogies carry about one bit of information, which is experimentally corroborated.
\end{abstract}
}

\maketitle

\IEEEdisplaynontitleabstractindextext

%
\IEEEpeerreviewmaketitle

\section{Introduction}\label{Sec:intro}
    
\IEEEPARstart{T}{exts} in natural language can be regarded as a kind of signal, but its non-numerical nature obviously prevents the use of most signal processing methods. As for signal prediction, however, the coding of words as continuous real-valued vectors seems to nicely circumvent this limitation. Indeed, Mikolov et al. \cite{Mikolov2013d} proposed a kind of prediction task where a machine was expected to solve analogies, thus predictions, such as the automatic guessing that the word like ``king'' is similar to ``man'' in the same sense as ``queen'' is similar to ``woman.'' Accordingly, it is expected that a good word embedding for words in English would associate vectors to words, say, ${\vec a}$, ${\vec b}$, ${\vec \alpha}$ and ${\vec \beta}$, for ``king'', ``queen'', ``man''  and ``woman'', respectively, in such a geometric configuration that the following approximation should hold:   
\begin{equation}
{\vec a} - {\vec \alpha} + {\vec \beta} \approx {\vec b}.
\end{equation}
This is referred to as the {\em vector offset method} (VOM) for solving word analogies, which has become a standard tool for semantic models in vector-spaces. 

Since \cite{Mikolov2013d} was published, word embeddings obtained with artificial neural networks became popular in applications ranging from document representation \cite{Kim2017} to sentiment analysis \cite{Cui2016}. Besides, word analogy became one of the tests frequently used to evaluate word embeddings (e.g. \cite{Lim2021, Choi2020,Sousa2020, Stein2019, Cao2019}), in spite of the VOM been criticized in its original formulation and results, to the point of suspicion regarding the actual relevance of analogies in the high accuracies frequently obtained. Remarkably, even the authors of \cite{Mikolov2013d} found ``somewhat surprisingly'' that analogy questions could be answered via the VOM, and to experimentally probe its prediction power, they prepared a list of 19,544 prediction tests, split into 8,869 semantic and 10,675 syntactic ones \cite{Mikolov2013e}, henceforth referred to as the Google Analogy Test Set (GATS). All tests followed the same format as $w_a$ is to $w_b$ as $w_\alpha$ is to $w_\beta$, where $w_x$ stands for the word associated to vector ${\vec x}$. 

Experiments done with the GATS yielded impressive results in terms of prediction accuracies with VOM, but it also attracted  some criticism, ranging from its statistical consistency  \cite{Linzen2016} and foundations \cite{Faruqui2016} to its optimization details \cite{Goldberg2014}, to the point that even the validity of the arithmetic analogy test has been questioned \cite{Levy2014}. {For instance, as pointed out in \cite{Linzen2016}, frustrating experimental results in analogy tests are observed if the instruction to discard input question words (before starting the output word search) is disregarded.}

This instruction is explicitly given in \cite{Mikolov2013d}, and the corresponding frustrating results suggest that vector 
\begin{equation}
\vec \phi_{\alpha,\beta} = {\vec \beta} - {\vec \alpha},
\end{equation}
\noindent which would encode the most valuable part of the analogy between pairs of words, may not play this expected role. On the other hand, proximity between similar words in the embedding is also expected to play a relevant role, as studied by Lund and Burgess \cite{Lund1996}, long before VOM was proposed. According to \cite{Lund1996}, neighbourhood surroundings are akin to a {\em semantic fields} by themselves, where similar representations tend to cluster words that can be substituted by each other in a context.  

In \cite{Fournier2020}, the relevance of vector $\vec \phi_{\alpha,\beta}$ for linguistic relations in word embeddings was studied through the pairing consistency score between $\vec \phi_{\alpha,\beta}$ and  
\begin{equation}
\vec \phi_{a,b} = {\vec b} - {\vec a},
\end{equation} 
\noindent where some pairs in analogy test sets were randomly shuffled, and Area Under the Curve values were subsequently computed to compare angle distributions with and without pair shuffling. 

Alternatively, the VOM formulation may be regarded as a word guessing game where $w_b$ is to be guessed from two separate hints, namely: that $w_b$ is similar to $w_a$, and that this similarity is analogous to that between $w_\alpha$ and  $w_\beta$. Because words are represented as real-valued vectors, this guessing game is also tantamount to a signal prediction where ${\vec a}$ plays the role of current signal instance, and $\vec \phi_{\alpha,\beta}$ should be a noisy direction for estimating $\vec b$ as a quantized version of prediction ${\vec p} = {\vec a} + \vec \phi_{\alpha,\beta}$. This signal prediction perspective further suggests that a simpler zero-order predictor would be ${\vec p}_1 = {\vec a}$, in which $\vec \phi_{\alpha,\beta}$ is not taken into account at all. This simpler predictor corresponds to the use of the first hint only, in the word guessing game, whereas the full predictor ${\vec p}_2 = {\vec a}+\vec \phi_{\alpha,\beta}$ uses the two hints.

In this work, we take this guessing game perspective to measure information content in popular analogy tests. In Section \ref{Sec:Expl}, the proposed approach is explained in terms of usual accuracy, whereas in Section \ref{Sec:Soft} an alternative accuracy measurement is introduced as a better estimator, for the purpose of this work. Experimental results are presented in Section \ref{Sec:results}, thus leading to the formulation of an analytical model for analogies in sparse word embeddings, in Section \ref{Sec:Model}. Conclusions are presented in Section \ref{Sec:DiscConc}.

\section{Measuring Information in Analogy Tests}
\label{Sec:Expl}

\begin{figure}[htb]
\centering{\includegraphics[width=90mm]{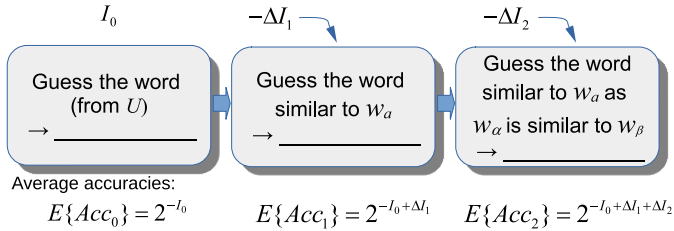}}
\caption{Word guessing in three levels of difficulty. From left to right: (1) Randomly guess a word in a set $U$ of $M$ words; (2) Guess a word $w_b$ whose corresponding vector is around $\vec a$ (3) Guess a word $w_b$ whose corresponding vector is around $\vec a$, toward $\vec \phi_{\alpha,\beta}$.}
\label{FigGame}
\end{figure}

\begin{figure}[htb]
\centering{\includegraphics[width=80mm]{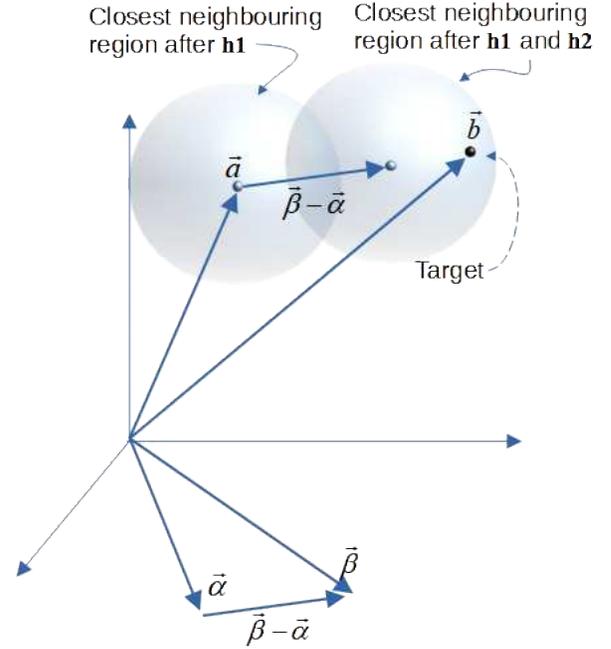}}
\caption{Illustration of how the two hints, {\bf h1} and {\bf h2}, modify the neighbouring region where the target vector is searched. Ideally, a good hint {\bf h2} is expected to move the search region toward the target, $\vec b$, which is associated to the correct answer $w_b$.}
\label{FigRegions}
\end{figure}

From an information perspective, each analogy test may be regarded as a guessing game where $w_b$ is a target symbol (word) to be guessed in a finite set $U$ of $M$ symbols. This perspective is illustrated in Fig. \ref{FigGame}, whereas Fig. \ref{FigRegions} illustrates a geometrical perspective of hints in analogy tests. Accordingly, two hints are provided as:
\begin{itemize}
\item[{\bf h1}] Similarity: $w_b$ is similar to $w_a$. Therefore, in the word embedding $\vec b$ is likely to be a near-neighbour of ${\vec p}_1 = \vec a$;
\item[{\bf h2}] Analogy: $w_\alpha$ relates to $w_\beta$ in the same manner as  $w_a$ relates to $w_b$, then $\vec b$ is likely to be a near neighbour of ${\vec p}_2 = {\vec a}+{\vec \beta}-{\vec \alpha}$.
\end{itemize}

{For the sake of a more intuitive presentation of ideas, the definition found in \cite{Mackay2003} for the Shannon information content of an outcome $x$ is used here, as:
\begin{equation}
I(x)=\log_2 \frac{1}{\Pr(X=x)}
\end{equation}
\noindent which turns out to be, in Shannon's original work \cite{Shannon1948}, the proposed ``measure of how much {\em  choice} is involved in the selection of the event or of how uncertain we are of the outcome''. The {\em information content} used in this work is a short for the {\em Shannon information content}, which is also defined for a random categoric outcome, such as the choice of a word. Therefore, Shannon's entropy is defined as the average information content. These connections to the Shannon's seminal work are further discussed in the Appendix. }

{Notice that, as in \cite{Montalvao2016}, accuracies can be alternatively thought in terms of entropy, through the concept of {\em effective cardinality}, as illustrated in the Appendix. Accordingly, any classification/detection problem, with a given accuracy, $A$, is analogous to another problem of finding a single target element in a {\em chimeric} set of $C=1/A$ equally likely ones, where $C$ is the effective cardinality of the {\em chimeric} set, and $H=\log_2(C)$ is the corresponding entropy, or its average information content, in bits. Effective cardinality is used here because it is assumed to intuitively reflect difficulty levels of a guessing game.}         

In the word guessing game illustrated in Fig. \ref{FigGame}, hints can be regarded as information content injections, $\Delta I_1 = I_0 - I_1$ and $\Delta I_2 = I1 - I2$, in bits, where only $\Delta I_2$ corresponds to the piece of information that gives name to (analogy) tests.  

Before any hint is provided, guessing the right word is worth $I_0$ bits. In natural language corpus, words are far from equally distributed, but there is no clear reason for assuming that a target word arbitrarily chosen for guessing purposes should follow any non-uniform distribution. Therefore, a conservative approach shall assume that $I_0=\log_2(M)$ bits, where $M$ is the cardinality of the set of all symbols, $U$, from which $w_b$ is randomly drawn. For instance, for a vocabulary of $M=400,000$ unique words in English, the guessing of $w_b$ without any hint has an information content of about 18.6 bits. Hint {\bf h1}, when provided, may have the power to shrink the search space to the near-neighbours of $w_a$, regardless the direction of these neighbours, which reduces the amount of incertitude about ${w_b}$ to $I_0 - \Delta I_1$ bits. Eventually, {\bf h2} may further reduce the incertitude about ${w_b}$ by moving the neighbourhood search in the direction ${\vec \phi}_{\alpha,\beta}$. This further reduction is expected even for a weak alignment between $\vec \phi_{\alpha,\beta}$ and $\vec \phi_{a,b}$, as studied in \cite{Fournier2020}, and finally the amount of incertitude about ${w_b}$ is reduced to $I_0 - \Delta I_1 - \Delta I_2$ bits.

Entropy measures reflect incertitudes about ${w_b}$, and are theoretically related to accuracies \cite{Montalvao2016}. Indeed, average accuracies can be expressed as 
\begin{equation}
E\{Acc_1\} = \frac{1}{2^{I_0-\Delta I_1}}
\end{equation}
\noindent and  
\begin{equation}
E\{Acc_2\} = \frac{1}{2^{I_0-\Delta I_1-\Delta I_2}},
\end{equation}
\noindent where $Acc_1$ and $Acc_2$ are random variables that model experimental accuracy instances, $acc_1$ and $acc_2$, of the guessing game, after {\bf h1} and {\bf h2} are provided, respectively. 


In turn, experimental accuracies, $acc_1$ and $acc_2$, can be used to estimate the amount of information carried out by each hint, as:
\begin{equation}\label{Eq:H1}
\widehat{\Delta I}_1 =  I_0 - \log_2(1/acc_1),\; \; acc_1 \neq 0 
\end{equation}
\noindent and
\begin{equation}\label{Eq:H2}
\widehat{\Delta I}_2 =   \log_2(1/acc_1) - \log_2(1/acc_2),\; \; acc_2 \neq 0
\end{equation}

\section{Soft accuracy}
\label{Sec:Soft}

\begin{figure}[htb]
\centering{\includegraphics[width=80mm]{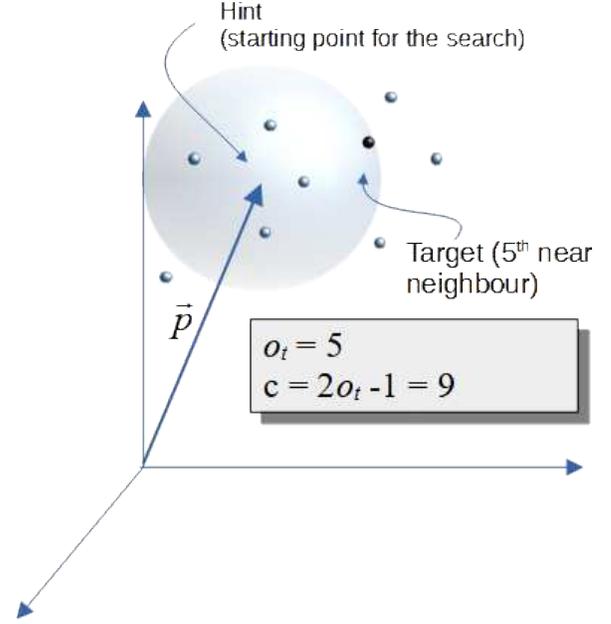}}
\caption{Illustration of the effective cardinality estimate for a single run of the guessing game, where the hint is that the target is around the point indicated by vector $\vec p$. However, the target is actually the 5$^{th}$ near neighbour from that point, thus $o_t=5$, and $c=9$ indicates that the target is the median of a set of 9 neighbours, taken as single independent instance of the search space cardinality.}
\label{FigEC}
\end{figure}

For finite sequences of a guessing game, accuracy estimates are typically spiky and slowly convergent, due to the binary nature of instances. To circumvent these problems, an improved accuracy estimator is proposed in this section, referred to as the {\em soft accuracy}, where instead of a (binary) hard decision/guess per experiment run, an effective cardinality is estimated from the position of the right answer. As illustrated in Fig. \ref{FigEC}, the instance estimate for effective cardinality, $c$, in a single experiment, is defined as             
\begin{equation}\label{Eq:EC}
c = 2o_t-1
\end{equation}  
\noindent {where $o_t$ is the sorted position of the target in ascending order of distance (or descending order or similarity), with regard to the hinted initial search position, i.e. prediction $\vec p$. In other words, $o_t$ stands for an instance of the random variable $O_t$, which models the sorted position of the target, thus it takes values from $\{ 1,2,\ldots,M \}$.} 

{This formulation stems from the belief that, in a huge set of possible answers, a small subset of $c$ answers around a given prediction are (almost) equally likely to be the target. Therefore, it is assumed that every observed instance of $o_t$ relates to a corresponding unobserved instance of $c$ according to $o_t = (c+1)/2$.}

Therefore, $c$ represents an instance of the random variable $C = 2O_t-1$, that models the effective cardinality of a search space around predictions, and takes values from $\{ 1,3,5,\ldots,2M-1 \}$. Accordingly, each independent instance of $c$ assumes that $o_t$ is one out of $2o_t-1$ neighbours of $\vec p$ that are equally likely to be the target,  thus $c$ roughly represents an average number of points around $\vec p$. In logarithmic scale, every independent guess, $g$, yields the following instance of information content of the right guess outcome (a hit): 
\begin{equation}\label{Eq:Pointh}
{\hat h}_g  = \log_2 (c) 
\end{equation}
\noindent {which is also the entropy of a hypothetical random variable uniformly distributed among $c$ values. Thus, over $G$ independent runs of the game, a ``naive'' entropy (i.e. average information content) estimate would be approximated as}
\begin{equation}\label{Eq:PointH}
{\hat H}_G  \approx \frac{\sum_{g=1}^{G} {\hat h}_g}{G} 
\end{equation}

Alas, this kind of entropy estimator has long been known to have a bias induced by the logarithm in Eq. \ref{Eq:Pointh}. Probably the simplest compensation for this kind of bias was proposed by G. Miller \cite{Miller1955}, in 1955. However, the entropy estimator in Eq. \ref{Eq:PointH} is not exactly the same studied by G. Miller, and a specific bias analysis is necessary here, where what we want is an estimate of $\log_2(E\{ C \})$, but we estimate $E\{ \log_2(C) \}$ instead. The above mentioned estimator bias is the resulting difference, $B$, formulated as   
\begin{equation}\label{EqBias_1}
E\{ \log_2(C) \} = \log_2(E\{ C \}) + B
\end{equation}

According to Eq. \ref{Eq:EC}, for target positions limited to $N$ near neighbours, the left side of Eq. \ref{EqBias_1} can be expanded as
\begin{equation} 
E\{ \log_2(C) \} = \sum_{n=1}^N \Pr(c_n) \log_2(2n-1).
\end{equation} 
\noindent If we further assume that all $N$ neighbours are equally likely to be the target, then it follows that 
\begin{equation}
E\{ \log_2(C) \} = \frac{1}{N} \sum_{n=1}^N \log_2(2n-1).
\end{equation}
\noindent By adding and subtracting $\log_2(N)$ in the right side of this equation, and after a few straightforward algebraic manipulation, we obtain
\begin{equation}\label{EqElogC_1}
E\{ \log_2(C) \} = \log_2(N) + \frac{1}{N} \sum_{n=1}^N \log_2\frac{2n-1}{N}
\end{equation}

On the other hand, for a uniform distribution, we have 
\begin{equation}\label{EqEC}
E\{ C \} = \frac{1}{N} \sum_{n=1}^N (2n-1) = N,
\end{equation}
\noindent and from  Eq. \ref{EqElogC_1} and  Eq. \ref{EqEC}, we obtain 
\begin{equation}\label{EqElogC_2}
E\{ \log_2(C) \} = \log_2(E\{ C \}) + \frac{1}{N} \sum_{n=1}^N \log_2\frac{2n-1}{N}
\end{equation}
Finally, by comparing Eq. \ref{EqElogC_2} to  Eq. \ref{EqBias_1} we conclude that 
\begin{equation}\label{EqBias_2}
B = \frac{1}{N} \sum_{n=1}^N \log_2\frac{2n-1}{N}
\end{equation}
  
Therefore, given $N$, and under the assumption that all $N$ neighbours are equally likely to be the target, a compensated entropy estimate is given by
\begin{equation}\label{EqCompensated}
\log_2(E\{ C \}) = {\hat H}_G - B
\end{equation}
\noindent and the soft accuracy can be obtained as   
\begin{equation}\label{Eq:SoftAcc}
acc = 2^{-{\hat H}_G + B}. 
\end{equation}

\begin{figure}[htb]
\centering{\includegraphics[width=80mm]{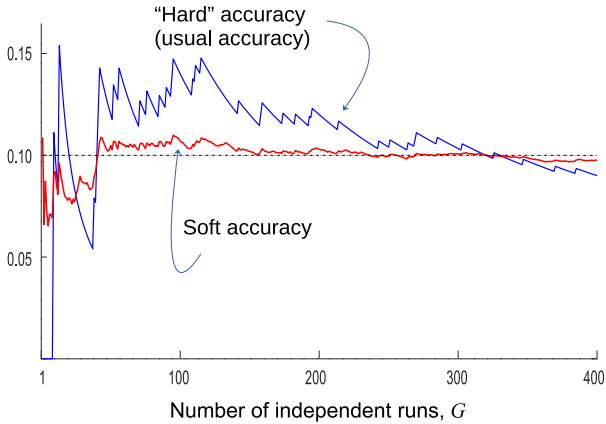}}
\caption{Illustration comparison between hard and soft accuracies for a guessing game with $N=10$, where hints are completely random (pseudo-hints). Both accuracies converge to $1/N = 0.1$.}
\label{FigHardvsSoft}
\end{figure}

An illustrative comparison between ``hard'' and soft accuracies is presented in Fig. \ref{FigHardvsSoft}, for a guessing game with $N=10$ and where a target, $o_t$, and a hint, $o_{hint}$, are randomly chosen in each independent run of the game. After $G$ such runs, the hard accuracy is given by the number of times $o_{hint}$ coincides with $o_t$ divided by $G$, whereas the soft accuracy is obtained as in Eq. \ref{Eq:SoftAcc}. Both accuracy estimates converge to $1/N = 0.1$, but the convergence of the soft one is faster and less susceptible to wide deviations. Therefore, in this work, rather than accuracy itself, entropy estimates as provided in Eq. \ref{EqCompensated} are to be used.  

{A possible drawback of this proposed soft accuracy would be that $N$ (the effective number of neighbours) must be known {\em a priori} to allow a proper bias compensation. Helpfully, the range of values that $B$ assumes is small. For instance, for $N$ in the range from 2 to 10,000, $B$ varies from  about $-0.20$ to about $-0.44$. In the experimental part of this work, we take advantage of it to assume a single heuristic value for (unknown) $N$, and then use the corresponding $B$ throughout all experiments. Indeed, in analogy tests, the target word is frequently either the first or the second near neighbour, sometimes the third, thus suggesting that the average number of near neighbours of predictions that are equally likely to be the target is not far from 2. Therefore, in experiments done in Section \ref{Sec:results}, the bias compensation value was arbitrarily set to $-0.25$.}

\section{Experiments} 
\label{Sec:results}

In what follows, Equations \ref{Eq:H1} and \ref{Eq:H2}, along with the soft accuracy defined in Eq. \ref{Eq:SoftAcc}, are articulated to build a pragmatic approach to measure the relevance of each hint that composes an analogy test, through the following kinds of experiments: 
\begin{itemize}
\item {\bf Single-hint experiment}: just word $w_a$ is taken into account, and the nearest neighbour of prediction $\vec p_1 = \vec a$, say $\vec g_1$, is taken as a tentative guess for $\vec b$. All vectors are normalized and proximity scores are obtained as inner products,
\begin{equation}
s_m = \langle {\vec p}_1, {\vec v}_m \rangle, \;  {\vec v}_m \neq {\vec a},
\end{equation}
\noindent where ${\vec v}_m$ is drawn from a set of $M-1$ vectors. In a test set with $G$ questions, for each question where $o_b=o_t$ is the order of the target score $s_b$, obtained for ${\vec v}_b$, $c_g$ is obtained according to Eq. \ref{Eq:EC}, representing the effective cardinality instance associated to the $g^{th}$ question/guess. Then an estimate ${\hat H}_G$ is accumulated as in Eq. \ref{Eq:PointH}, thus allowing the computation of a soft accuracy instance, $acc_1$, as in Eq. \ref{Eq:SoftAcc}. 

\item {\bf Two-hints experiment}: a prediction is given by $\vec p_2 = \vec a + {\vec \beta}-{\vec \alpha}$. Again, all vectors are normalized and proximity scores are obtained as  
\begin{equation} 
s_m = \langle{\vec p}_2, {\vec v}_m \rangle, \;  {\vec v}_m  \notin \{ {\vec a}, {\vec \alpha}, {\vec \beta} \},
\end{equation} 
\noindent where ${\vec v}_m$ is now drawn from a set of $M-3$ vectors, and the corresponding soft accuracy instance, $acc_2$, is obtained. 
\end{itemize}

The first set of experiments was performed with a pre-trained set of $M = 400,000$ words in English, encoded with GloVe \cite{Pennington2014} as 300-D vectors. The corresponding embedding of this publicly available dataset is referred to as the {\em Wikipedia 2014 $+$ Gigaword 5}.  As for the tests, we used both the GATS \cite{Mikolov2013e}, in Table \ref{Tab:WG_Google}, and the Bigger Analogy Test Set (BATS)\cite{Gladkova2016}, in Table \ref{Tab:WG_BATS}.
\begin{table}[htb]
\centering
\caption{Soft accuracies (in \%) and information content variations (in bits) for the GATS.}\label{Tab:WG_Google}
\begin{tabular}{|c|c|c|c|c|}
\hline 
Subsets & Soft & Soft & $\widehat{\Delta I}_1$ & $\widehat{\Delta I}_2$ \\
	    & accuracy & accuracy &  &  \\
	 	& (single-hint) & (two-hints) &  &  \\
\hline
capital-common- & & & & \\
-countries & 25.5 & 80.2 & 16.6 & 1.7 \\
\hline
capital-world & 27.5 & 81.0 & 16.7 & 1.6 \\
\hline
city-in-state & 4.5 & 28.2 & 14.1 & 2.6 \\
\hline
currency & 0.1 & 0.5 & 7.9 & 3.0 \\
\hline
family & 30.1 & 68.5 & 16.9 & 1.2 \\
\hline
\hline
gram1-adjective- & & & & \\
-to-adverb & 3.2 & 5.2 & 13.6 & 0.7 \\
\hline
gram2-opposite & 1.0 & 2.5 & 12.0 & 1.3 \\
\hline
gram3-comparative & 17.4 & 68.9 & 16.1 & 2.0 \\
\hline
gram4-superlative & 1.3 & 46.6 & 12.4 & 5.1 \\
\hline
gram5-present- & &  &  &  \\
-participle & 27.6 & 44.6 & 16.8 & 0.7 \\
\hline
gram6-nationality- & & & & \\
-adjective & 29.5 & 68.8 & 16.8 & 1.2 \\
\hline
gram7-past-tense & 21.6 & 37.5 & 16.4 & 0.8 \\
\hline
gram8-plural & 42.4 & 59.5 & 17.4 & 0.5 \\
\hline
gram9-plural-verbs & 7.3 & 36.1 & 14.8 & 2.3 \\
\hline 
\end{tabular} 
\end{table}
\begin{table}[htb]
\centering
\caption{Soft accuracies (in \%) and information content variations (in bits) for the BATS.}\label{Tab:WG_BATS} 
\begin{tabular}{|c|c|c|c|c|}
\hline 
Subsets & Soft & Soft & $\widehat{\Delta I}_1$ & $\widehat{\Delta I}_2$ \\
	    & accuracy & accuracy &  &  \\
	 	& (single-hint) & (two-hints) &  &  \\
\hline \hline 
Inflectional morph. & 7.5  & 27.3 & 14.9 & 1.9  \\
\hline
Derivational morph. & 0.2  & 0.4 & 9.6 & 0.9 \\
\hline \hline 
Encyclopedic semant. & 0.4 & 2.0 & 10.6  & 2.4 \\
\hline
Lexicographic semant. & 0.7 & 0.3 & 11.4 & -1.0  \\
\hline 
\end{tabular} 
\end{table}

As expected, all results are consistent with the conclusions already published in many former works (e.g. \cite{Linzen2016}, \cite{Faruqui2016}, \cite{Goldberg2014}, \cite{Levy2014} and \cite{Fournier2020}), that the analogy itself is not as important as accuracies based on both hints may suggest. Additionally, presented values in bits for $\widehat{\Delta I}_1$ and $\widehat{\Delta I}_2$ provide a suitable quantitative perception of it. For instance, while in Table \ref{Tab:WG_Google} the GATS have analogies with average information content of about 1.8 bits, which is already small, as compared to the average 15.3 bits for $\widehat{\Delta I}_1$, results in Table \ref{Tab:WG_BATS} corroborate the expectation that BATS is a more challenging set of analogies, yielding even a loss of information content of about 1 bit, for the  {\em Lexicographic semant.} subset. This loss is expressed as a negative value that means that {\bf h2} is a disturbing noise --- rather than a true hint --- that doubles the average search space for target words. {It is worth noticing that this negative value is not an information content, which cannot be negative, by definition. A simple illustration of this kind of negative result is presented in the Appendix.}

\section{Simple model for analogies in word embeddings} \label{Sec:Model}

Experimental values of $\widehat{\Delta I}_2$ in both tables \ref{Tab:WG_Google} and \ref{Tab:WG_BATS} yield an average value of about 1.6 bits, which is much less than any value of $\widehat{\Delta I}_1$. To afford some understanding of such disparity, even a coarse model for the word embedding can be of help, as follows: let all $M$ words to appear in pairs $(\vec a_m, \vec b_m)$, $m= 1, 2, \ldots M/2$, so that they almost satisfy $\vec b_m \approx {\vec a}_m+{\vec \phi}_{\alpha,\beta}$, where a constant ${\vec \phi}_{\alpha,\beta}$ encodes a single unique analogy relationship for all pairs of word in this coarse model. Besides, the embedding is assumed to be sparse (as actual word embeddings \cite{Fournier2020}), {therefore most directions point out toward space regions where points representing words are far apart from each other.} 

\begin{figure}[htb]
\centering{\includegraphics[width=80mm]{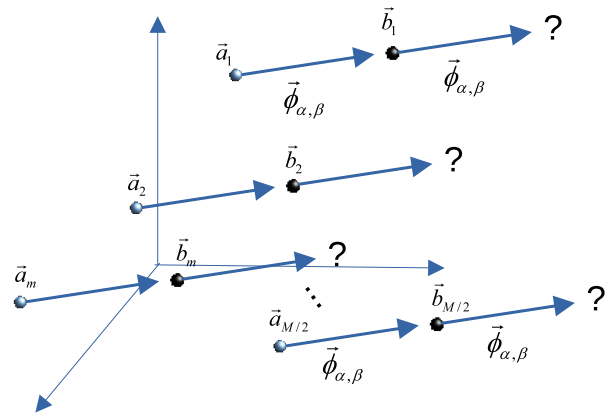}}
\caption{Illustration of how start-words and end-words yield qualitatively different predictions when added to vector ${\vec \phi}_{\alpha,\beta}$. Start-words, represented by $\vec a$ vectors, yield prediction close to  $\vec b$ vectors, representing end-words, whereas end-words yield predictions most probably far from other vectors. As a result, when guessing candidates are sorted according to their respective distances to respective predictions, the target $\vec b_t$ is expected to be found among the first $M/2$ near neighbours.}
\label{FigModel}
\end{figure}

Before any hint is provided, an observer cannot know whether a word is of kind $a$ (start-word) or $b$ (end-word). When only hint {\bf h2} is provided, direction ${\vec \phi}_{\alpha,\beta}$ is given, but the start-word is not. Consequently,  ${\vec \phi}_{\alpha,\beta}$ must be added to all words, as illustrated in Fig. \ref{FigModel}, because the guesser is unaware of which are the start-words, thus yielding $M$ predictions. However, only for the $M/2$ actual start-words the approximation 
\begin{equation}
\vec b_m \approx {\vec a}_m+{\vec \phi}_{\alpha,\beta}
\end{equation} 
\noindent is warranted, which is likely to place the actual target among the $M/2$ actual end-words, whereas the embedding sparseness is likely to lead most predictions from end-words,
\begin{equation}
{\vec p} = {\vec b}_m+{\vec \phi}_{\alpha,\beta},
\end{equation}
\noindent {to regions of the embedding where even the first near neighbour is far away. Therefore one should expect to find the target $\vec b_t$ among the $M/2$ closest near neighbours (NN) of all $M$ predictions.}

Again, due to the approximation
\begin{equation}
\vec b_m \approx {\vec a}_m+{\vec \phi}_{\alpha,\beta},
\end{equation}
\noindent  {the $M/2$ closest near neighbours are expected to be almost equally likely to be the tentative guess. According to equations from Eq. \ref{EqEC} to \ref{Eq:SoftAcc}, if we replace the $N$ equally probable symbols with $M/2$, the expected soft accuracy is indeed $2/M$, for the search space is reduced, on average, to  $M/2$ (almost) equally likely predictions, and }
\begin{equation} 
I_0 - \log_2(M/2) = 1 \; bits.
\end{equation}  



{Therefore, according to this coarse model, {\bf h2} is expected to carry about 1 bit of information, for it halves the search space, when {\bf h1} is not provided and the word embedding is sparse. However, the effect of {\bf h2} after {\bf h1} is less evident. In practice, results such as $\widehat{\Delta I}_2 = 0.7$, for the subset {\em gram1-adjective-to-adverb}, in Table \ref{Tab:WG_Google}, seems to suggest that the effective search space is reduced to more than half that spotted by {\bf h1}, whereas $\widehat{\Delta I}_2 = 5.1$, for the subset {\em gram4-superlative}, suggests a much stronger reduction of this search space to less than 3\% of the words spotted by {\bf h1}. Besides, from the same perspective, the negative result in Table \ref{Tab:WG_BATS}, $\widehat{\Delta I}_2 = -1.0$, for the {\em Lexicographic semant.} subset suggests that hint {\bf h2} is misleading the guesser, doubling in size the search space already spotted by {\bf h1}. A simple illustration for how is it possible is presented in the Appendix. }     

\section{Conclusion}
\label{Sec:DiscConc}

An approach for measuring the information content of hints in analogy tests was proposed. For conciseness, all experiments were done with a pre-trained set of GloVe 300-D vectors, whose performance in analogy tests is considered representative of most state-of-the-art embeddings. The test sets were either the Google Analogy Test Set, or the Bigger Analogy Test Set. All experiments corroborate the general perception noticed in publications since 2016  \cite{Linzen2016, Faruqui2016, Goldberg2014, Levy2014, Linzen2016}, that the analogy hint is much less relevant to the test performances than word proximity, even if word vectors do capture linguistic regularities. This was confirmed in all but one subset of tests, where the analogy hint had a negative effect, suggesting that sometimes analogy hints may even play the role of a disturbing noise, or a false hint for the guessing game. Indeed, the negative value of $\widehat{\Delta I}_2$, in Table \ref{Tab:WG_BATS}, for the last subset of tests, means that analogy hints hindered the target word search.

{Results, when regarded solely in terms of accuracy, may suggest that the analogy hint is strongly relevant, at least for the purpose of accuracy gain in tests. However true, this is a fallacious result inasmuch as accuracy falsely amplifies the actual importance of {\bf h2}. For instance, a result such as $\widehat{\Delta I}_1 = 16.8$, for the subset {\em gram6-nationality-adjective}, in Table \ref{Tab:WG_Google}, indicates that {\bf h1} reduced the search space from $M=400,000$ to a set of spotted words with effective cardinality of about $4$ words, yielding an accuracy of about 29.5\%. After {\bf h1}, {\bf h2} has a very modest effect of further reducing the search space to a set whose effective cardinality is about $2$ words, but the accuracy impressively increases to 68.8\%. Indeed, the accuracy gain is amplified by the fact that it is inversely proportional to the effective size of the search space, already mostly reduced by the effect of {\bf h1}.}

In \cite{Fournier2020}, another measure for the presence of linguistic relations in word embeddings was used, the Pairing Consistency Score (PCS), which quantifies the degree to which offset vectors ${\vec \phi}_{a,b}$ and ${\vec \phi}_{\alpha,\beta}$ are parallel above chance. Deviations between angle distributions were measured with Area Under the Curve (AUC) values. That is to say that AUC values in the range from 0.5 to 1.0 were used to evaluate the relevance of analogies in each set of experiments, and PCS values greater than 0.5 were taken as indicators that ${\vec \phi}_{a,b}$ and ${\vec \phi}_{\alpha,\beta}$ were parallel above chance. From the information content perspective, within-category shuffling does break analogies, so that the corresponding pseudo hint {\bf h2} should carry null information content, which indeed corresponds to what suited experiments in \cite{Fournier2020}. Unfortunately, the resulting deviations of PCS from 0.5, which are proposed there as measures of linguistic relations relevance, are not easily translatable into accuracy gains due to these same linguistic relations, otherwise it would be an interesting matter for direct comparison with the approach proposed here.

{As for indirect comparisons, however, from results presented in \cite{Fournier2020}, the pairing consistency score is more deviated from 0.5 for the {\em Inflectional} subset of BATS, whereas $\widehat{\Delta I}_2 = 1.9$, in Table \ref{Tab:WG_BATS}, is the second highest estimated value of ${\Delta I}_2$ for BATS, thus corroborating the relevance of analogy in that subset. By contrast, PCS is larger for subset {\em Derivational} than for the {\em Encyclopedic} one, while the opposite is observed in terms of $\widehat{\Delta I}_2$, in Table \ref{Tab:WG_BATS}. Moreover, although PCS is close to 0.5 for tests with the {\em Lexicographic} subset, it is still greater than 0.5, which means that offsets $\vec \phi_{a,b}$ and $\vec \phi_{\alpha,\beta}$ are parallel above chance, on average, whereas in Table \ref{Tab:WG_BATS} we find a negative variation of information content due to hint {\bf h2}. }

{These results are not inconsistent. In particular, for the {\em Lexicographic} subset, PCS value suggests a non-random parallelism between  $\vec \phi_{a,b}$ and $\vec \phi_{\alpha,\beta}$, in that subset, whereas the negative $\widehat{\Delta I}_2$ suggests that $\vec \phi_{\alpha,\beta}$ is moving the search toward a neighbourhood with more target candidates than the prediction neighbourhood after {\bf h1} alone. Indeed, because  $\widehat{\Delta I}_2 = -1$ bit, we may even infer that $\vec p_2 = \vec a + \vec \phi_{\alpha,\beta}$ spots a portion of the search space with effective cardinality twice as larger as that spotted by $\vec p_1 = \vec a $, on average. Therefore, the two results do not contradict each other, for although $\widehat{\Delta I}_2$ also depends on the parallelism between  $\vec \phi_{a,b}$ and $\vec \phi_{\alpha,\beta}$, as PCS, it also depends on the effective cardinality of spotted portions of the search space.} 
                                 
Therefore, the proposed approach allows a new complementary way of measuring the actual relevance of analogy hints, as compared to proximity hints. Besides, the vector offset method is just one kind of signal prediction model that can be used in word embeddings, and the proposed approach for information content estimation has the potential to be a useful tool in further studies regarding signal prediction in embeddings.


\appendix[A brief illustration of Effective Cardinality and Information Content]

Two main concepts used in this work are {\em effective cardinality} and {\em information content}. To illustrate these concepts, we consider a game where a (loaded) six-sided dice is thrown and a random variable $X$ is associated to the corresponding outcomes, where $\Pr(X=1)=1/6$, $\Pr(X=2)=1/24$, $\Pr(X=3)=1/3$, $\Pr(X=4)=1/12$, $\Pr(X=5)=1/3$  and $\Pr(X=6)=1/24$.

The Shannon Information Content of an outcome $x$ is defined in \cite{Mackay2003} in bits as 
\[
I(x)=\log_2 \frac{1}{\Pr(X=x)}
\]
\noindent which turns out to be, in Shannon's original work (\cite{Shannon1948}, Section 7), the proposed {\em measure of how much ``choice'' is involved in the selection of the event or of how uncertain we are of the outcome}. The Information Content in this work is a short for the Shannon Information Content, which is also defined for a random categoric outcome, such as the choice of a word. 

Before any hint is provided, a guesser has no reason to prefer any outcome, then he randomly chooses one out of the six faces at random. Therefore, the chances of each face being chosen by the guesser is $1/6$, and whatever his choice may be, the average accuracy of this guesser can be easily computed as being $1/6$, in spite of the non-uniform  probability distribution of the loaded dice. Therefore, $I_0 = \log_2(6)$ bits quantifies the missing information for the guesser, even though the probability distribution of the dice is not uniform. It comes from the fact that the unbalanced distribution is not known by the guesser.    

Then, a first hint, {\bf h1}, is provided, according to which outcomes from the loaded dice have an expected value  around 3 and 4, which is true, for $E\{X\}=3.5$. Suppose the guesser reacts to {\bf h1} by updating his model of the dice distribution to $\Pr(X=1) \approx 1/8$, $\Pr(X=2) \approx 1/8$, $\Pr(X=3) \approx 1/4$, $\Pr(X=4) \approx 1/4$, $\Pr(X=5) \approx 1/8$  and $\Pr(X=6) \approx 1/8$. Indeed, this distribution roughly reflects what the guesser just learned from {\bf h1}, and now, to maximize his chances of a hit, he chooses either face 3 or face 4. Therefore, the average accuracy of this guesser increases to $5/24$, then $2^{-I_1}=5/24$, or  $I_1 = \log_2(24/5) \approx 2.26$ bits, and ${\Delta I}_1 \approx 0.32$ bits.

At this point, the concept of {\em effective cardinality} can be conveniently introduced as follows: from the viewpoint of the  guesser, in spite of the actual probability distribution of the loaded dice, the first problem was that of finding a single right answer in a set of $C_0 = 2^{I_0} = 6$ equally probable ones. Clearly enough, $C_0$ is the actual cardinality of this set. After {\bf h1} was provided, this problem was modified to finding a single right answer in a (chimerical) set of $C_1 = 2^{I_1} = 4.8$ equally probable ones. In this case, $C_1$ is not an actual set cardinality, but it plays the role of an {\em effective cardinality} that intuitively reflects the difficulty level of the guessing game, in this context. More generally, effective cardinality can be a useful dual for the entropy concept itself, as explained in \cite{Montalvao2016} and references therein.         
       
Finally, to illustrate why ${\Delta I}_2$ can be negative, consider that a second hint, {\bf h2}, was provided, according to which outcomes from the loaded dice are more likely to be 4, 5 or 6, which is false! Now, not knowing that the hint is wrong, the guesser may update his former model to also include {\bf h2}, as $\Pr(X=1) \approx 1/13$, $\Pr(X=2) \approx 1/13$, $\Pr(X=3) \approx 3/13$, $\Pr(X=4) \approx 4/13$, $\Pr(X=5) \approx 2/13$  and $\Pr(X=6) \approx 2/13$. 

From the guesser perspective, To maximize his chances of a hit, in accordance to this updated model, his guess should be always face 4. However, because {\bf h2} is a false hint, the guesser accuracy in lessened to $1/12$, or  $I_2 = \log_2(12) \approx 3.58$ bits, and ${\Delta I}_2 \approx -1.32$ bits. Alternatively, one may consider that the guessing game was hardened by the misleading hint {\bf h2}, from searching the right answer in a set of $C_1 = 4.8$ equally probable ones, to searching in a bigger virtual set of $C_2 = 12$ also equally likely answers. It is noteworthy that it is even harder than guessing the outcome of a fair dice with 6 faces, thus with cardinality $C_0$. In other words, the original effective cardinality was expanded as a consequence of a false hint.   

\end{document}